\documentclass[runningheads]{llncs}
\usepackage[T1]{fontenc}
\usepackage{graphicx}
\usepackage{booktabs}
\usepackage{amsmath}
\usepackage{multirow}
\begin{document}
\title{Phrase-grounded Fact-checking for Automatically Generated Chest X-ray Reports}
%
%\titlerunning{Abbreviated paper title}
% If the paper title is too long for the running head, you can set
% an abbreviated paper title here
%
\author{Razi Mahmood\inst{1}, Diego Machado-Reyes\inst{1}, Joy Wu \inst{2,3},
Parisa Kaviani\inst{4}, \\
Ken C.L. Wong\inst{2}, Niharika D'Souza\inst{2}, Mannudeep Kalra\inst{4}, Ge Wang\inst{1}, \\Pingkun Yan\inst{1}, Tanveer Syeda-Mahmood\inst{2,3}}
\authorrunning{R. Mahmood et al.}
% First names are abbreviated in the running head.
% If there are more than two authors, 'et al.' is used.
%
\institute{Rensselaer Polytechnic Institute, NY, USA ,\and
IBM Research, Almaden, CA, USA \and
Stanford University, CA, USA, \and 
Massachusetts General Hospital (MGH), Boston, USA\\
\email{mahmor@rpi.edu, stf@us.ibm.com}\\
%\url{http://www.springer.com/gp/computer-science/lncs} \and
%ABC Institute, Rupert-Karls-University Heidelberg, Heidelberg, Germany\\
%\email{\{abc,lncs\}@uni-heidelberg.de}
}
\maketitle              % typeset the header of the contribution
\begin{abstract}
With the emergence of large-scale vision language models (VLM), it is now possible to produce realistic-looking radiology reports for chest X-ray images. However, their clinical translation has been hampered by the factual errors and hallucinations in the produced descriptions during inference. In this paper, we present a novel phrase-grounded fact-checking model (FC model) that detects errors in findings and their indicated locations in automatically generated chest radiology reports. 
Specifically, we simulate the errors in reports through a large synthetic dataset derived by perturbing findings and their locations in ground truth reports to form real and fake findings-location pairs with images.  A new multi-label cross-modal contrastive regression network is then trained on this dataset. We present results demonstrating the robustness of our method in terms of accuracy of finding veracity prediction and localization on multiple X-ray datasets. We also show its effectiveness for error detection in reports of SOTA report generators on multiple datasets achieving a concordance correlation coefficient of 0.997 with ground truth-based verification, thus pointing to its utility during clinical inference in radiology workflows.
\end{abstract}
\section{Introduction}
Preliminary radiology reports generated by automated report generation models are valuable in emergency room settings, where radiologists may not be immediately available and rapid interpretation is required\cite{syeda-mahmood2020}. Current methods of report generation are predominantly based on vision language models (VLM)\cite{maira,Pang2023,Gao2024,Ranjit2023,Ramesh2022} which still suffer from hallucinations and factual errors that limit their clinical applicability\cite{gale2018producing}. Strategies to correct such models exist such as direct policy optimization (DPO)\cite{rafailov2023direct,Hardy2024,Passi2022,zhou2023analyzing} or proximal policy optimization (PPO)\cite{zheng2023secrets} with reward models\cite{ziegler2019finetuning} directly tap into the generative decoder parameters to compute hallucination risk scores \cite{Hardy2024}. However, they are applicable during training or fine-tuning stages. Methods of fact-checking at inference time exist but they often consult external knowledge \cite{Passi2022,nieman,midas} to spot the factual errors which are unsuitable for radiology reports as the report needs to be specific to the patient-image. Similarly, methods that use generic large language models (LLMs) as judges to verify the radiology report text \cite{maira,Ramesh2022,Zhang2024} are not suitable either, since they themselves have hallucinations, and may not corroborate their deductions with the patient-specific image. 

Thus, there is a need to develop an independent fact-checking method for the clinical inference phase to bootstrap radiology report generation. Realizing this, we had earlier attempted to build a simple fact-checking model using a pre-trained vision-language model (CLIP) and a binary SVM to classify sentences as real or fake in automated reports\cite{mlmi2023}.  However, being based on full sentences, it was sensitive to writing styles in reports. Further, it used a frozen encoder and did not offer any explanation of the errors nor perform a phrasal grounding of the findings in images. 
%Very little attention has been paid so far to post-training fact-checking of radiology reports with most work being focused on report generation~\cite{maira,Pang2023,Gao2024,Ranjit2023}. 
\begin{figure}[t]
\centering
  % \centerline{\includegraphics[width=\textwidth]{revisedfigs/overviewcombined.png}}

%\centerline{\includegraphics[width=\textwidth]{revisedfigs/new_merged_fig1-2.png}}
%\centerline{\includegraphics[width=0.5\textwidth]{revisedfigs/fctrain2.png}\includegraphics[width=0.5\textwidth]{revisedfigs/fcinfer2.png}}
%\centerline{\includegraphics[height=0.45\textheight]{revisedfigs/all3vertical.png}}
\centerline{\includegraphics[width=5in,height=3.5in]{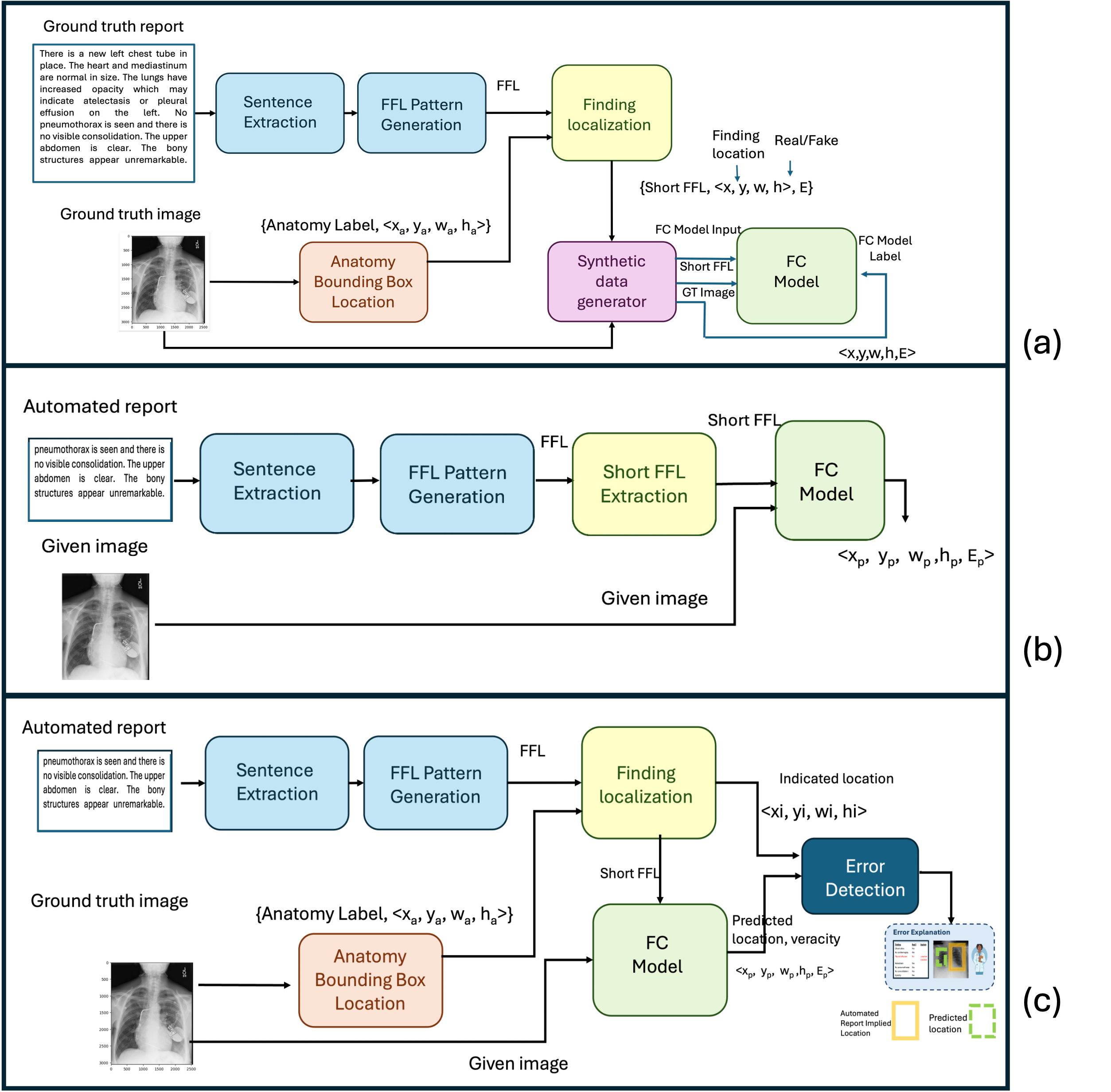}}
  \caption{
  % \py{Figs. 1 and 2 can be merged. Embed the example in Fig. 1 to Fig. 2. Also rearrange the boxes in Fig. 2 to better utilize the space.} 
  %\razi{Merged figures 1-2} 
  Illustration of workflow for FC model training (a) , inference (b), and error detection and explanation (c) using common modules.} 
  %\py{This figure is too small. Can you try to merge the two into one and use different colors to denote training and inference input/output? The main modules are shared anyway.}}
  \label{overallapproach}
\end{figure}

In this paper, we introduce an innovative method of fact-checking chest X-ray radiology reports during clinical inference with the following novel contributions. First, we derive a large synthetic dataset of over 27 million images paired with real and fake findings to simulate errors in reports through perturbation of their identities and location descriptions in ground truth reports. This dataset is now being contributed to open source. We also develop a new multi-label contrastive regression model for fact-checking (FC model) that is trained to discriminate and anatomically ground the real and fake findings. Through extensive testing, we show that the FC model can detect radiology report errors with a concordance correlation coefficient of 0.997 with ground truth-based verification making it a potential surrogate for ground truth during clinical inference.  
\vspace{-0.1in}
\section{Method}
\label{prepare}
Our overall approach to factual error detection in automated radiology reports consists of training a fact-checking model (Figure~\ref{overallapproach}a), using it in inference mode on automated reports to record predicted findings and their locations (Figure~\ref{overallapproach}b), and recording the deviations of implied findings from automated reports from predictions as shown in Figure~\ref{overallapproach}c. As can be seen from Figure~\ref{overallapproach},  the workflows use common pre-processing modules of sentence extraction, finding extraction and anatomy localization, but are fed different image-text pairs during training and inference. Specifically, the training workflow depicted in Figure~\ref{overallapproach}a involves: (i) finding localization, (ii) synthetic data generation and (iii) FC model training. Step (i) extracts anatomical locations (L) from ground truth images (I), findings (F) from their reports, and collates to generate bounding boxes $<x,y,w,h>$ for findings. In step (ii) synthetic perturbations are applied to generate real/fake pairs $<F,I,x,y,w,h,E>$ where $E$ is the veracity label. In step (iii) the FC model is trained using $<F,I>$ as input and $<x,y,w,h,E>$ as output. During inference, the FC model is given findings extracted from automated reports and their image as input, to predict the output  $<x_p,y_p,w_p,h_p,E_p>$ where $<x_p,y_p,w_p,h_p>$ is the bounding box and Ep is the predicted real/fake label as shown in Figure~\ref{overallapproach}b. Finally, the error detection and quantification workflow shown in Figure~\ref{overallapproach}c recovers the indicated location from automated report and compares it to the predicted finding and location using an error measure that results in a visual explanation.
 %Specifically,  the training workflow (shown through red lines) uses a synthetic data generator to perturb the location and identities of findings derived from ground truth reports and images. The inference workflow (shown in green) uses the given image and the automatically generated report to predict the veracity of the finding and its location. The error detection and quantification workflow (shown in orange) recovers the indicated location from automated report and compares to the predicted finding and location using an error measure that results in a visual explanation. We now describe the approach in detail.
\vspace{-0.2in}
\subsection{Training dataset generation}
\vspace*{-0.5\baselineskip}
For training data generation, we use prior work on finding extraction\cite{syeda-mahmood2020} and anatomical region detection \cite{wu2020a,syeda-mahmood2020a} as pre-processing. To make our fact-checking approach agnostic to sentence writing styles, we abstract the described findings in sentences into a simplified structured form called FFL (fine-grained finding labels) using the method described in \cite{syeda-mahmood2020} and as illustrated in Table~\ref{ffl}. Each finding is normalized to a standard vocabulary (\textit{e.g.}, pulmonary vasculature engorged -> vascular congestion) using a  comprehensive clinician-curated chest X-ray lexicon of 101,088 distinct FFL \cite{wu2020a,syeda-mahmood2020a} which are sufficient to capture the variety seen in automatically generated reports. The FFL extraction algorithm reported in \cite{syeda-mahmood2020} had a 97\% accuracy and was seen as sufficient for our pre-processing. In addition to finding descriptions, we also use the anatomical location algorithm described in \cite{Wu2020-isbi2020,Wu2021} to locate bounding boxes in any frontal chest X-ray image for 36 anatomical regions cataloged in the chest X-ray lexicon \cite{wu2020a,syeda-mahmood2020a}. Its accuracy was previously assessed at 0.896 precision and 0.881 recall, and was used to generate the Chest ImaGenome benchmark dataset\cite{Wu2021}. 
%For example in a sentence "Pleural vasculature is not engorged and the patient has moderate pulmonary edema on the right" would be reduced to two structured FFLs as "anatomicalfinding|no|vascular congestion|lung" and "anatomicalfinding|yes|pumonary edema|right lung".
\begin{table}[t]
\centering
\caption{Illustration of FFL.}\label{ffl}
\scalebox{0.95}{
\begin{tabular}{l|c}
\hline
\textbf{Sentence} & \textbf{Simplified FFL} \\
\hline
Pleural vasculature is not engorged & anatomicalfinding | no | vascular congestion | lung\\
and the patient has moderate  & 
anatomicalfinding | yes | pumonary edema|right lung\\
pulmonary edema on the right \\
\hline
\end{tabular}}
\end{table}

\begin{table}[t]
  \centering
  \caption{Illustration of synthetic perturbations to produce the training dataset for the FC model. Only the core finding in column 2 for simplicity. \\
  'E' : (0=non-existent finding, 1=existing finding) 
  }
  \label{synthetic}
  \begin{tabular}{l|l|l}
    \toprule
    Synthetic Perturbation & Generated Finding & Label ($<$xy,w,h,E$>$) \\
    \hline
   Original 	& yes$|$edema 	& $<0.14, 0.13, 0.72, 0.56, 1>$\\
Reversal 	& no$|$edema 	& $<0,0,0,0,0>$\\
Relocate 	& yes$|$edema 	& $<0.85,0.74, 0.10, 0.21, 0>$\\
Relocate 	& yes$|$edema 	& $<0.90,0.70,0.10,0.20,0>$\\
Substitution 	& yes$|$lung cyst 	& $<0.02,0.48, 0.10, 0.14, 0>$\\
    \bottomrule
  \end{tabular}
\end{table}
Let $<I,R>$ be the sample set of ground truth image-report pairs in a gold dataset $D$. Let $F=\{F_{j}\}$ be the total list of possible findings in chest X-ray datasets. %Each report-image pair $R_{i}$ will contain a variable number of findings, an existing multi-label set of sample $D_{i}\in D=<I_{i},R_{i}>$ 
The set of real Finding-location (FL) pairs extracted by the pre-processing per sample $D_i  = <I_i, R_i> \in D $
% $D_{i}\in D=<I_{i},R_{i}>$
can be denoted by $FL_{iReal} =\{fl_{ij}\}=\{<f_{ij},l_{ij}>\}$ where:
\begin{equation}
    f_{ij} =<T_{ij}|N_{ij}|C_{ij}>,
    l_{ij}=<x_{ij},y_{ij},w_{ij},h_{ij}>.
\end{equation}
Here $f_{ij}\in F_{iReal} $ is the jth real finding in report $R_{i}$  and $l_{ij}$ is the bounding box for the finding $f_{ij}$ in image $I_{i}$ starting at $(x_{ij},y_{ij})$ of width $w_{ij}$ and height $h_{ij}$ in normalized coordinates ranging from 0 to 1.

Let $L_{j}=\{l_{ij}\}$ be the list of all normalized locations accumulated across all images of $D$ for a finding $F_{j}$. With normalized coordinates, and since we pick among the valid finding locations, any synthetic location generated for $F_{j}$ will be valid for some image in the dataset.   %Randomly drawing from this set ensures that a synthetic location generated for $F_{j}$ is a valid location for some image in the dataset. 

The errors found in generated reports are known to include false predictions, incorrect finding locations, omissions, or incorrect severity assessments\cite{Yu2023}. We focus on the first two errors so that given a real finding $f_{ij}$ at location $l_{ij}$ for a sample $D_{i}$, we create 3 variants to reflect (a) reversal of polarity  (b) relocation of the finding (c) and substitution with appropriate relocation as 
%\begin{equation}
$FL_{iFake}=\{<\overline{fl_{ij}}, fl_{ik},fl_{mn}> \}$,
%\end{equation}
where $\overline{fl_{ij}}$ is the reversed finding, $fl_{ik}$ is finding $f_{ij}$ relocated to a random new position $l_{k}\in L_{j}$, and $fl_{mj}$ is obtained by randomly substituting finding $f_{j}$ with $f_{m}\not\in F_{i}$ at location $l_{n}\in L_{m}$ taking care to avoid repeats and contradictions. Table~\ref{synthetic} shows synthetic perturbations created from an original finding "yes$|$edema" based on the operations above.

\vspace{-0.1in}
\subsection{Building the FC model}
\begin{figure}[t]
\centering
  \centerline{\includegraphics[width=5.0in]{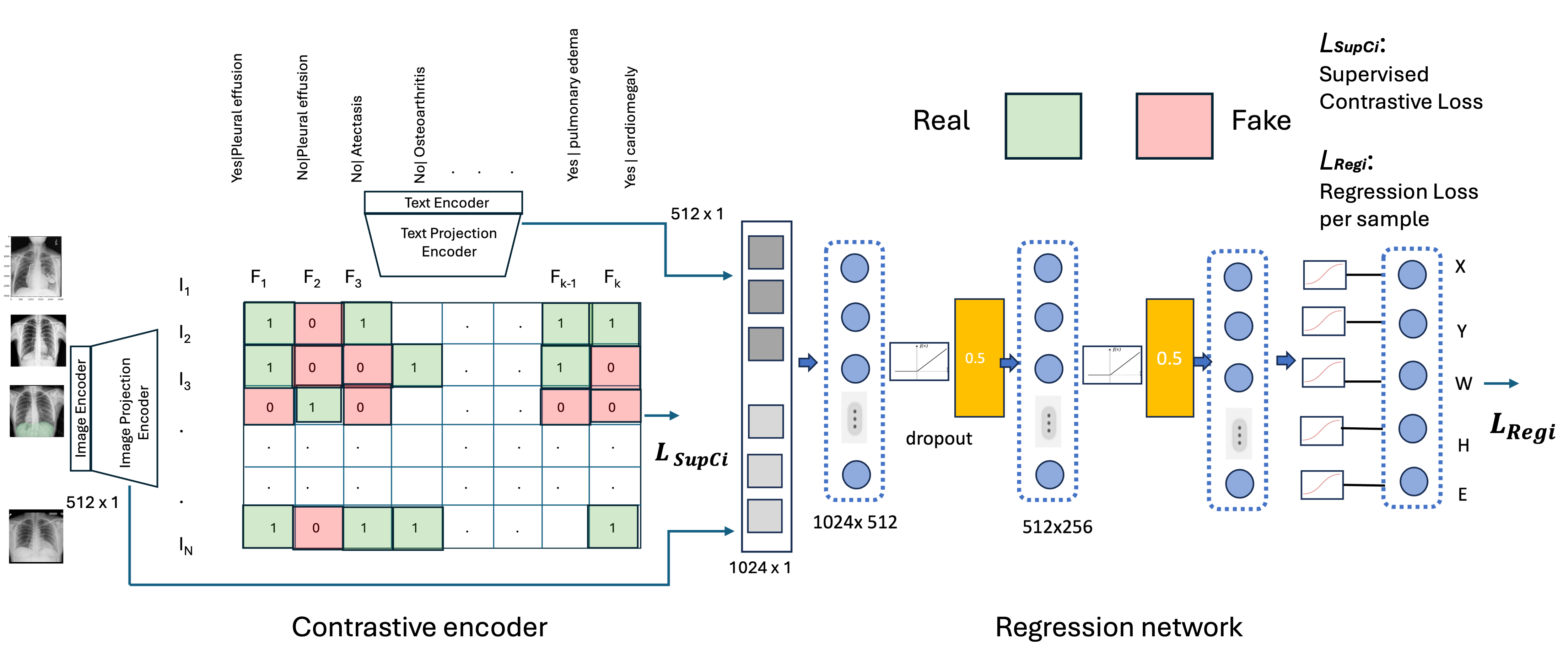}}
  \caption{Illustration of the architecture of our FC model. The real FFL are taken as positive and the fake FFL as negative in the contrastive formulation. }
  \label{fcarchitecture}
\end{figure}

\vspace*{-0.5\baselineskip}
The end-to-end architecture of the FC model is illustrated in Figure~\ref{fcarchitecture}. We use the encodings of images and FFL text to learn a joint embedding space that is designed to separate the real FFL labels from fake labels using supervised contrastive learning\cite{khosla2020supervised}. The embeddings from real and fake text-image pairs are then concatenated to learn an inner regression network to predict both the location and veracity of the finding. 

Let $z_{i}$ be the vision projection encoder output, and let $z_{f_{ij}}$ be the text encodings of findings for each sample $D_{i}=(I_{i},F_{i})$ where $f_{ij}\in F_{i}=F_{iReal}\cup F_{iFake}$ are the real and fake labels per sample. We define a multi-label cross-modal supervised contrastive loss per sample as:
\begin{equation}
\mathcal{L}_{SupC_{i}}=\frac{-1}{|F_{iReal}|} \sum_{f_{ij}\in F_{iReal}} log\frac{e^{s_if_{ij}/\tau}}{\sum_{a_{ik}\in F_{iFake}} e^{s_{ia_{ik}}/\tau}}
  \label{multilabelloss}
\end{equation}
\noindent where $s_{if_{ij}}=z_{i}\cdot z_{f_{ij}}$ is the pairwise cosine similarity between image and textual embedding vectors from the real findings $f_{ij}\in F_{iReal}$, and $s_{ia_{ik}}=z_{i}\cdot z_{a_{ik}}$ is with the fake findings where $a_{ik}\in F_{iFake}$. The overall loss is obtained by averaging across all the samples in the batch. Here $\tau$ is the temperature parameter. This formulation results in a non-diagonal similarity matrix as shown in Figure~\ref{fcarchitecture} and differs significantly from existing VLM contrastive encoders based on CLIP who all assume a diagonal similarity matrix\cite{clip,mlmi2023,Ramesh2022} and are self-supervised. It also differs from supervised contrastive learning which was previously unimodal and used for image classification from augmented version of images treated as positive samples\cite{khosla2020supervised}.  {\em To our knowledge, the supervised contrastive learning formulation has not been used to develop vision-language encoders with real and fake labels}. 

Next, the inner regression network takes the projected joint embeddings $T_{ijReal}=[z_{i}|z_{f_{ij}}]$ of  image $I_{i}$ paired with real finding label $f_{ij}\in F_{iReal}$ or fake labels $T_{ijFake}=[z{i}|z_{a_{ik}}]$ where $a_{ik}\in F_{iFake}$ and the corresponding supervision label $Y_{g}=<Y_{1g},Y_{2g}>$ where $Y_{1g}=<x,y,w,h>$ and $Y_{2g}=E=1$ for the real finding and 0 otherwise. Using $Y_{p}=<Y_{1p},Y_{2p}>$ as the prediction from the network, we can express the regression loss per sample as 
\begin{eqnarray}
\mathcal{L}_{Reg_{i}} =\underbrace{{\vert{Y_{1p}-Y_{1g}\vert}}}_{\mathcal{L}_{1}(Y_{1p}, Y_{1g})}+ \underbrace{\frac{\vert{Y_{1p} \cap Y_{1g} \vert}}{\vert{Y_{1p} \cup Y_{1g}}\vert} - \frac{\vert C_{Y_{1p},Y_{1g}} \backslash {Y_{1p} \cap Y_{1g}}\vert}{\vert{C_{Y_{1p},Y_{1g}}}\vert} }_{\mathcal{L}_{\text{giou}}(Y_{1p}, Y_{1g}) } \nonumber \\
+ \underbrace{{\vert{Y_{1p}-Y_{1g}}\vert}^{2}}_{\mathcal{L}_{\text{mse}}(Y_{1p}, Y_{1g})}  - \underbrace{[{Y_{2g}\textbf{log}(Y_{2p})} + {(1-Y_{2g})\textbf{log}(1- Y_{2p})}]}_{\mathcal{L}_{BCE}(Y_{2p},Y_{2g})}
\label{regressionloss}
\end{eqnarray}
where $C_{Y_{1p},Y_{1g}}$ is the convex hull of the bounding boxes defined by $Y_{1p}$ and $Y_{1g}$.

The loss function reflects the dual attributes being optimized, namely, the location and the veracity of the finding.  The L1 loss and generalized IOU loss have previously been used for regression\cite{chen23MedRPG}. However, since in our case, the negative findings have bounding box coordinates as $
<0,0,0,0>$ which poses a problem for generalized IOU when the prediction error is small. For this reason, and to ensure smooth convergence, we added the mean square penalty. Finally, for the veracity indicator variable $E$, we use the binary cross entropy loss. 

{\noindent\bf\underline{Implementation details}}: We used a chest X-ray pre-trained CLIP encoder (151,277,313 parameters)
\cite{Ramesh2022} and retained its  image encoder (ViT-B/32) and text encoder (masked self-attention Transformer). The joint embedding projection layers of CLIP (768x512 for image and 512x512 for text) were, however, fresh-trained using the new supervised contrastive formulation derived from real-fake labels. The regression network (657,413 parameters) consisted two linear layers, two drop out layers with RELU for intermediate layers and separate sigmoidal functions for producing the output regression vectors as shown in Figure~\ref{fcarchitecture}. To train this network in an end-to-end fashion, the losses defined in Equations~\ref{multilabelloss} and ~\ref{regressionloss} were applied at the respective heads shown in Figure~\ref{fcarchitecture}. The FC model was trained for 100 epochs using the AdamW optimizer on an NVIDIA A100 GPU with 40GB of memory and a batch size of 32. The cosine annealing learning rate scheduler was used with the maximum learning rate of 1e-5 and 50 steps for warm up. 

\begin{table*}[t]
\caption{Details of datasets used in experiments. Here CImagenomeS stands for Chest ImagGenome silver dataset. }\label{datasets}
\centering
\scalebox{0.95}{
\begin{tabular}{llllll}
\hline
\multirow{2}{*}{\textbf{Dataset}} & \textbf{Patients} & \textbf{ Images} & \textbf{ Findings} & \textbf{Regions} & \textbf{Real/Synth.} \\
& \textbf{Train/Val/Test} &  &  & &
\\
\hline
CImagenomeS\cite{Wu2021} & 44,133/6274/12,538 & 243,311 & 49 & 922,295 & 1.616M/27.047M\\
CImaGenomeG\cite{Wu2021} & 288/33/69 & 461 & 35 & 5,477 & 4,063/23,463\\
MS-CXR\cite{mimic-4} & 478/54/114 & 925 & 8 & 2,254 & 2,247/24,338 \\
ChestXray8\cite{ChestXay8} & 457/51/109 & 880 & 8 & 1,571 & 1,571/10,137 \\
VinDr-CXR\cite{vindrcxr} & 9,450/1,050/2,250 & 15,000 & 23 & 69,052 & 47,973/132,632  \\
\hline
\end{tabular}}
%\caption{Details of datasets used in experiments.}
\end{table*}
\vspace{-0.1in}
\subsection{Error detection using the FC model}
\label{fcinference}
\vspace*{-0.5\baselineskip}
%To detect errors in generated report during deployment, %the same pre-processing done for ground truth reports is repeated on the generated reports and the given image, to derive indicated bounding box locations of anatomical regions of the reported findings 
%as shown in  Figure~\ref{overallapproach},  
The FFL extracted from an automated report and the image are used by the FC model to predict the veracity of the finding label and its location as shown in Figure~\ref{overallapproach}b. To quantify the error, we use a phrase-grounded error measure called the FC score\cite{isbi2025} which was shown to outperform other evaluation measures such as Radgraph F1, SBERT, and BLEU score.  Specifically, we calculate the error detected by FC model as $RQ(A,P)= 1-\mbox{FCScore(A,P)}$ as :
\begin{equation}
RQ(A,P)==1-\frac{1}{2}(\frac{|E_{pj}=1|}{\sum_{E_{pj}\in E_{p}}E_{pj}}+\frac{1}{|L_{p}|}{\sum_{j}
\frac{\vert{L_{Aj} \cap L_{pj} \vert}}{2\vert{L_{Aj} \cup L_{pj}}\vert}})
\label{fcscore}
\end{equation}
Here  $E_{pj}$ is a predicted veracity for an indicated label $F_{Aj}\in F_{A}$ in the automated report, and  $L_{Aj}, L_{pj}$ are the indicated locations from automated reports and the predicted locations from the FC model respectively computed as shown in Figure~\ref{overallapproach}c.

\begin{table}[t!]
\caption{This table illustrates multiple aspects of the FC model evaluation. The FC model performance under different ablation architecture configurations across multiple datasets are rows in the first 4 rows. The last two rows show comparison of our FC model's phrasal grounding and real/fake classification performance against SOTA methods. }\label{comparison}
\begin{tabular}{c|c|c|c|c|c|c|c|c}
\hline
\multirow{2}{*}{\textbf{Method}} &  \multicolumn{2}{|c|}{\textbf{CImaGenomeG}} &\multicolumn{2}{|c|}{\textbf{MS-CXR}} &\multicolumn{2}{|c|}{\textbf{ChestX-ray8}} &\multicolumn{2}{|c}{\textbf{VinDR-CXR}}\\
& Accuracy & MIOU & Accuracy & MIOU & Accuracy & MIOU & Accuracy & MIOU \\
\hline
{\bf FCRegComb.}& {\bf 0.92} &  {\bf 0.54} & {\bf 0.94} & {\bf 0.53} & {\bf 0.92} & {\bf 0.57} & {\bf 0.90} & {\bf 0.49}\\
FCRegBCE & 0.88 &  0.49 & 0.92 & 0.46 & 0.90 & 0.53 & 0.88 & 0.45\\
FCRegDual & 0.87 &  0.51 & 0.89 & 0.49 & 0.87 & 0.51 & 0.86 & 0.47\\
FCRegSep  & 0.89 &  0.38 & 0.89 & 0.39 & {\bf 0.92} & 0.42 & 0.89 & 0.37\\
\hline
Med-RPG\cite{chen23MedRPG} & -  & 0.23 & - & 0.32 & - & 0.28 & - & 0.38\\
Maira-2\cite{maira} & - & 0.39 & - & 0.48 &- & 0.51 & - & 0.42\\
R/F Model\cite{mlmi2023} & 0.84 & - & 0.78 & - & 0.81 & - & 0.83 & -\\
\hline
\end{tabular}
\end{table}

\begin{figure}[t]
\centering
    \centerline{\includegraphics[width=5in]{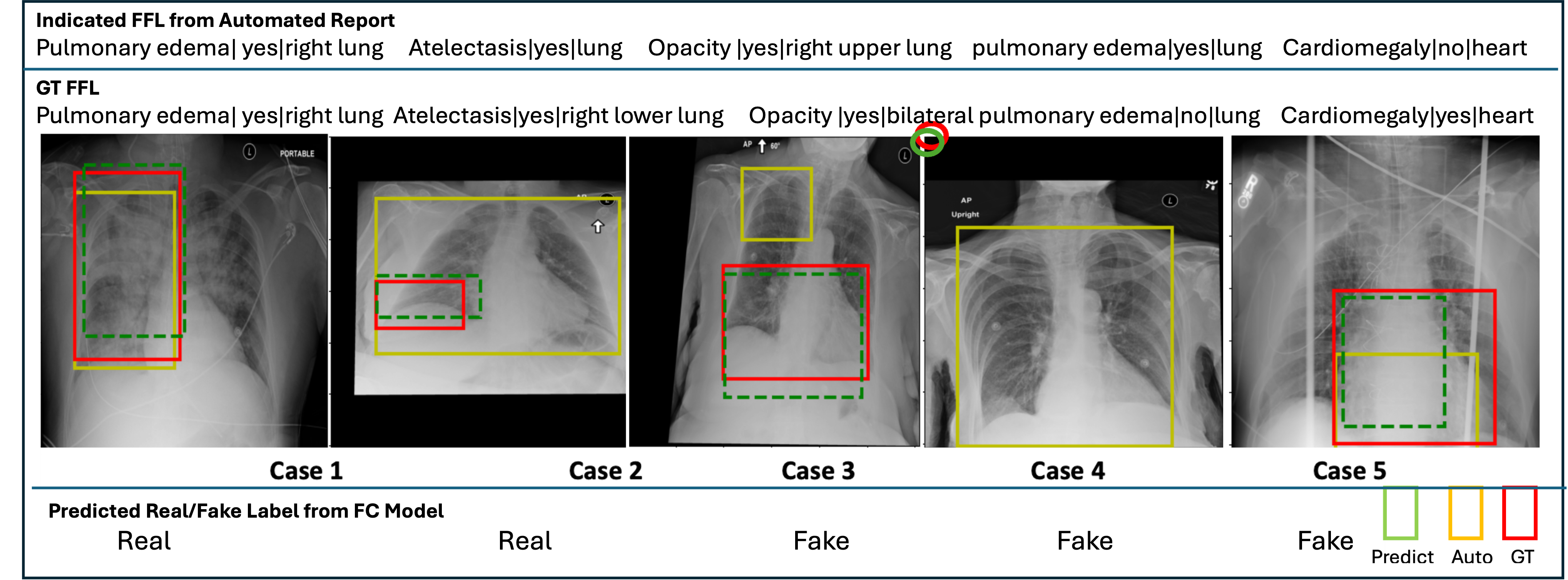}}
  \caption{Illustration of error detection and localization for 5 sentences from reports generated by X-rayGPT\cite{xraygpt}. The legend for bounding boxes: predicted finding location: Green, indicated finding: orange, ground truth finding: red. } 
  \label{exampleresults}
\end{figure}

\begin{table}[t!]
\caption{Illustrating the effectiveness of the FC model in assessing errors in generated reports. High concordance can be seen between error detection using ground truth (A,G) and error detection  using FC model (A,P) in all cases.}\label{assessmentscore}
\begin{tabular}{c|c|c|c|c|c|c|c|c}
\hline
\multirow{2}{*}{\textbf{Report Generator}} &  \multicolumn{2}{|c|}{\textbf{CImaGenomeG}} &\multicolumn{2}{|c|}{\textbf{MS-CXR}} &\multicolumn{2}{|c|}{\textbf{ChestX-ray8}} &\multicolumn{2}{|c}{\textbf{VinDR-CXR}}\\
& \multicolumn{2}{|c|}{$RQ$}  & \multicolumn{2}{|c|}{$RQ$}  & \multicolumn{2}{|c|}{$RQ$} &  \multicolumn{2}{|c|}{$RQ$}\\
\hline
& (A,P) & (A,G)& (A,P) & (A,G) & (A,P) & (A,G) &  (A,P) & (A,G) \\
\hline
RGRG\cite{rgrg}  & 0.541 &  0.537 & 0.329 & 0.308 & 0.305 & 0.298 & 0.549 & 0.537\\
XrayGPT\cite{xraygpt} & 0.622 & 0.626 & 0.388 & 0.391 & 0.377 & 0.355 & 0.618 & 0.609\\
GPT4-inhouse & 0.658 &  0.653 & 0.433 & 0.426 & 0.399 & 0.408 & 0.636 & 0.630\\
R2GenGPT\cite{R2GenGPT} & 0.587 &  0.585 & 0.377 & 0.374 & 0.346 & 0.333 & 0.581 & 0.579\\
CV2DistillGPT2\cite{CV2DistillGPT2} & 0.576 & 0.573 & 0.439 & 0.433 & 0.427 & 0.420 & 0.588 & 0.6\\
CheXRepair\cite{Ramesh2022} &  0.744 & 0.733 & 0.466 & 0.461 & 0.439 & 0.432 & 0.709 & 0.714\\
Maira-2\cite{maira} &  0.619 & 0.633 & 0.423 & 0.425 & 0.412 & 0.419 & 0.578 & 0.569\\
\hline
\end{tabular}
\end{table}
%\vspace{-0.2in}

\section{Results}
\vspace*{-0.5\baselineskip}
We conducted several experiments using chest X-ray datasets with location and finding annotations shown in Table~\ref{datasets}. Of these, Chest ImaGenome gold (CImagenomeG)\cite{Wu2021} dataset was set aside  for error detection evaluation as it had a complete set of ground truth reports, clinician-verified findings and their locations\cite{Wu2021}.  The training partitions of the rest of the datasets were used for the generation of the synthetic dataset yielding over 27 million samples as shown in Table~\ref{datasets}.  

{\noindent\bf\underline{Automated report generators evaluated:}} We selected several SOTA report generators whose code was freely available as shown in Table~\ref{assessmentscore}. All report generators were given the same prompt, and automated reports collected for the 439 images of the (CImagenomeG) were retained for error analysis.

\noindent{\bf\underline{Real/Fake classification performance:}} We evaluated the accuracy of FC model's in FFL veracity prediction using the test partitions of the datasets shown in Table~\ref{datasets}.  The model consistently yielded an accuracy over 90\% for real/fake classification, as shown in Table~\ref{comparison}. By using 10 fold cross-validation in the generation of the (70-10-20) splits for the datasets, the average accuracy of the test sets lay in the range 0.92 ± 0.12.

{\noindent\bf\underline{Anatomical grounding performance}}: Figure~\ref{exampleresults} illustrates sample explainable error detection by the FC model on XrayGPT-generated reports\cite{xraygpt}). By comparing the bounding box locations and predicted labels to ground truth FFLs, we observed that the FC model correctly flags errors and localizes findings with greater overlap with ground truth. In fact, the mean IOU with the ground truth bounding boxes ranged from 0.49-0.57 as shown in Table~\ref{comparison} (rows 1-4), across various model architectures.

 \noindent{\bf\underline{Comparison to related methods:}} With no prior work on fact-checking with phrasal grounding for chest X-ray reports, we compared to the nearest methods that either do phrasal grounding  (MED-RPG\cite{chen23MedRPG},Maira-2\cite{maira}) or real/fake classification (the R/F Model from \cite{mlmi2023}).  The results are shown in Table~\ref{comparison} in the last three rows recording the relevant numbers for a  regressor or classifier respectively. In comparison to pure phrase grounding or real/fake classification only, our method predicts both veracity and location of findings, and outperforms these methods across all the datasets. 
 
\noindent{\bf\underline{Ablation studies:}} We conducted ablation studies using 4 different architectures, namely, (a) end-to-end training as shown in Figure~\ref{fcarchitecture} (FCRegComb), (b) replacing supervised contrastive loss with BCE loss (FCRegBCE), (c) using  a generic pre-built CLIP encoder with regressor (FCRegSep), and (d) using a dual head regressor with separate loss functions for regression and classification (FCRegDual). The results of real/fake classification and phrasal grounding shown in Table~\ref{comparison} indicate that combining the contrastive encoder with the regressor in an end-to-end fashion gave the best performance.

{\noindent\bf\underline{Fact-checking report performance:}} Fact-checking involves computing the error between the indicated (A) and predicted FL pairs (P) as $RQ(A,P)$ and comparing it to $RQ(A,G)$  of indicated FL pairs with the ground truth $G$. These results are summarized in Table~\ref{assessmentscore} averaged across all images for each report generator tested.   As can be seen, the $RQ(A,P)$  has good correlation with $RQ(A,G)$ and the overall concordance correlation coefficient\cite{Lin1989} with the ground truth at 0.997. In comparison, using the real/fake classifier model\cite{mlmi2023}, the concordance correlation coefficient was lower at 0.831 since the location errors could not be verified. These results show the potential of fact-checking models for error detection during inference in clinical workflows even when no ground truth is available. 

\vspace{-0.1in}
\section{Conclusions}
\vspace{-0.1in}
In this paper, we have presented a new fact-checking model for chest X-ray reports that detects errors in findings and their reported locations. The model has a high concordance coefficient with the ground truth for error estimation pointing to its utility as a surrogate for ground truth during inference. Future work on the FC model will address findings omitted from reports, and explore ways of incorporating it during the training phases to further improve report generators. 
%
% the environments 'definition', 'lemma', 'proposition', 'corollary',
% 'remark', and 'example' are defined in the LLNCS documentclass as well.
%

\begin{credits}
\subsubsection{\discintname}
The authors have no competing interests. 
\end{credits}
%
% ---- Bibliography ----
%
% BibTeX users should specify bibliography style 'splncs04'.
% References will then be sorted and formatted in the correct style.
%
% \bibliographystyle{splncs04}
% \bibliography{mybibliography}
%

{\small
\bibliographystyle{splncs04}
\bibliography{phrasegrounded}
}
\end{document}